\documentclass[preprint,12pt]{elsarticle}

\usepackage{amssymb}
\usepackage{amsmath}

\usepackage{algorithmic}
\usepackage{algorithm}
\usepackage{array}
\usepackage[caption=false,font=normalsize,labelfont=sf,textfont=sf]{subfig}
\usepackage{textcomp}
\usepackage{stfloats}
\usepackage{url}
\usepackage{booktabs}
\usepackage{verbatim}
\usepackage{graphicx}
\usepackage{tabularx}
\usepackage{color}
\usepackage{xcolor}
\usepackage{multirow}
\usepackage{colortbl}
\usepackage{orcidlink} 
\usepackage{soul}
\usepackage{ulem}
\newcommand{\vcenteredbox}[1]{\begingroup
\setbox0=\hbox{#1}\parbox{\wd0}{\box0}\endgroup}

\journal{Knowledge-Based Systems}

\begin{document}

\begin{frontmatter}



\title{Vision-based 3D Semantic Scene Completion
via Capture Dynamic Representations}


\author{Meng Wang} 
\ead{willem@hnu.edu.cn}

\author{Fan Wu}
\ead{wufan@hnu.edu.cn}

\author{Yunchuan Qin} 
\ead{qinyunchuan@hnu.edu.cn}

\author{Ruihui Li\corref{cor1}} 
\ead{liruihui@hnu.edu.cn}

\author{Zhuo Tang} 
\ead{ztang@hnu.edu.cn}

\author{Kenli Li} 
\ead{lkl@hnu.edu.cn}

\cortext[cor1]{Corresponding author.}
\affiliation{organization={College of Computer Science and Electronic Engineering, Hunan University},
            city={Changsha},
            postcode={410082}, 
            country={China}}

\begin{abstract}
The vision-based semantic scene completion task aims to predict dense geometric and semantic 3D scene representations from 2D images. However, the presence of dynamic objects in the scene seriously affects the accuracy of the model inferring 3D structures from 2D images. Existing methods simply stack multiple frames of image input to increase dense scene semantic information, but ignore the fact that dynamic objects and non-texture areas violate multi-view consistency and matching reliability.
To address these issues, we propose a novel method, CDScene: Vision-based 3D Semantic Scene Completion via Capturing Dynamic Representations.
First, we leverage a large multi-modal model to extract 2D explicit semantics and align them into 3D space.
Second, we exploit the characteristics of monocular and stereo depth to decouple scene information into dynamic and static features. The dynamic features contain structural relationships around dynamic objects, and the static features contain dense contextual spatial information.
Finally, we design a dynamic-static adaptive fusion module to effectively extract and aggregate complementary features, achieving robust and accurate semantic scene completion in autonomous driving scenarios.
Extensive experimental results on the SemanticKITTI, SSCBench-KITTI360, and SemanticKITTI-C datasets demonstrate the superiority and robustness of CDScene over existing state-of-the-art methods.
\end{abstract}





\end{frontmatter}


\section{Introduction}

With the rapid development of autonomous driving, the field of 3D scene understanding has encountered new challenges. Autonomous vehicles~\cite{shen2023flowformer} rely on accurate environmental perception to ensure safe navigation and effective obstacle avoidance. However, obtaining precise and complete 3D representations of real-world environments remains challenging due to factors such as limited sensor resolution, restricted fields of view, and occlusions that lead to incomplete observations. Semantic scene completion (SSC)~\cite{song2017semantic,yan2021sparse} method to simultaneously infer the complete scene geometry and semantic information from sparse observations.

Existing SSC solutions~\cite{guo2018view,roldao2020lmscnet} primarily rely on input RGB images and corresponding 3D data to predict volume occupancy and semantic labels. However, the dependence on 3D data often requires specialized and expensive depth sensors, limiting the broader application of SSC algorithms. Recently, many researchers~\cite{cao2022monoscene, li2023voxformer,li2023stereoscene,jiang2024symphonize,tits2024instance,mixssc,wang2025vlscene} have begun exploring pure vision-based solutions to address 3D SSC.

However, the presence of dynamic objects in the scene seriously affects the accuracy of the model inferring 3D structures from 2D images. Most of the aforementioned methods estimate depth using a single image frame. Monocular depth estimation methods~\cite{bhat2021adabins, yin2022towards,yan2023dsc} rely on semantic understanding of the scene and perspective projection cues, making them more robust to textureless regions and dynamic objects, while not depending on camera pose. 
Multi-frame methods~\cite{gu2020cascade,yao2018mvsnet,Yu_2020_fastmvsnet,chuah2022semantic} estimate depth based on the assumption that, given accurate depth, camera calibration, and camera pose, pixels should appear similar across views. However, the accuracy and robustness of multi-view methods are heavily dependent on the geometric configuration of the camera and the corresponding pixel matching between views. First, when multi-view methods encounter ubiquitous dynamic objects (such as moving cars and pedestrians), dynamic areas can cause cost volume degradation due to violations of multi-view consistency. Second, for untextured or reflective surfaces, multi-view matching becomes unreliable.
\begin{figure*}
    \centering
    \includegraphics[width=\linewidth]{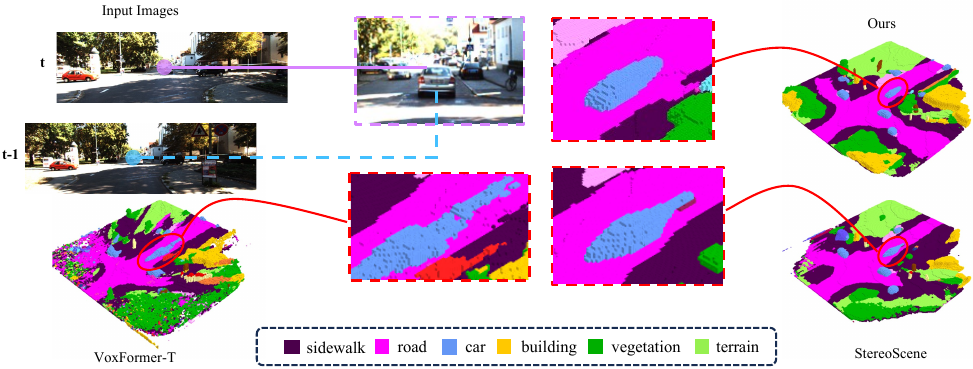}
    \caption{Given a 2D image from a camera, our method demonstrates accurate modeling of dynamic objects in the scene. There are no long traces in the semantic representation of the dynamic vehicle in the red box, and the geometry of the entire scene is more detailed.}
    \label{fig:fig1}
\end{figure*}

To address the above challenges, we propose a novel framework, CDScene: Vision-based 3D Semantic Scene Completion via Capture Dynamic Representations. The key insight is to leverage explicit semantic modeling of objects in the scene and decouple scene information into dynamic and static features. {We propose to use monocular depth to focus on dynamic regions (e.g., moving car and person), where appearance-based depth cues remain effective, and stereo depth to guide static region reconstruction, offering precise spatial information. This modality-specific decomposition allows us better to capture the scene's underlying structure and motion properties.} Our method effectively models dynamic objects within the scene, as shown in Fig.~\ref{fig:fig1}, where there are no long traces in the semantic expression of dynamic vehicles, and the geometry of the entire scene is represented in a more detailed dense expression.

Specifically, CDScene first uses a large multi-modal model(LMMs) to extract object semantic features, which are sampled into voxel space to provide semantic modeling information required for 3D scene understanding. Second, we analyze the idiosyncrasies of monocular and stereo depth in driving scenes and decouple features into two branches: the dynamic branch and the static branch. The dynamic branch learns structural relationships around dynamic objects, while the static branch captures dense contextual spatial properties in the scene. Finally, we design a dynamic-static adaptive fusion module to integrate features from the two branches in voxel space, combining robust features around dynamic objects with precise geometric features of static regions.
To evaluate the performance of CDScene, we conduct extensive experiments on the SemanticKITTI~\cite{behley2019semantickitti}, SSCBench-KITTI360~\cite{li2023sscbench}, and SemanticKITTI-C datasets. The results surpass those of existing methods by a significant margin, establishing state-of-the-art performance. The main contributions of our work are summarized as follows:
\begin{itemize}
    \item We introduce the LMMs provide the explicit semantic modeling information required for 3D scene understanding.
    \item We decouple dynamic and static features, where the dynamic branch captures effective structural relationships around dynamic objects, and the static branch is used to extract dense context geometric properties in static regions.
    \item We design a dynamic-static adaptive fusion module to effectively extract and aggregate complementary features, achieving robust and accurate semantic scene completion in autonomous driving scenarios.
    \item The proposed CDScene model achieves state-of-the-art results on the SemanticKITTI, SSCBench-KITTI360 and SemanticKITTI-C datasets, surpassing the latest approaches to demonstrate the superiority and robustness.
\end{itemize}
\section{Related Work}

\subsection{3D Semantic Scene Completion}
The SSC task has garnered significant interest due to its crucial role in semantic occupancy prediction for autonomous driving. The introduction of the extensive outdoor benchmark SemanticKITTI~\cite{behley2019semantickitti} has led to the emergence of various outdoor SSC techniques. Existing outdoor methods are primarily categorized into lidar-based and camera-based methods based on their input modality. Lidar-based approaches prioritize the use of LiDAR for precise 3D semantic occupancy prediction.
SSCNet~\cite{song2017semantic} is an innovative study that first presents the idea of semantic scene completion, emphasizing the simultaneous inference of geometry and semantics from incomplete visual information. 
UDNet~\cite{zou2021udnet} employs a unified 3D U-Net structure to make predictions based on a grid derived from LiDAR points. 
Subsequently, LMSCNet~\cite{roldao2020lmscnet} introduced 2D CNN for feature encoding, and SGCNet~\cite{zhang2018efficient} used spatial group convolution to improve efficiency. 
Further research has recognized the inherently three-dimensional nature of the semantic scene completion task. This understanding has led to several studies exploiting 3D inputs such as depth information, occupancy meshes, and point clouds, exploiting their rich geometric cues~\cite{rist2021semantic, cai2021semantic, li2019rgbd,wan2024bi}. 
Furthermore, this study delves into the connection between semantic segmentation and scene completion. 
For instance, JS3C-Net~\cite{yan2021sparse} and SSA-SC~\cite{yang2021ssasc} have developed semantic segmentation networks to aid in semantic scene completion. 
Additionally, SSC-RS~\cite{mei2023sscrs} has proposed a multi-branch network for the hierarchical integration of semantic and geometric features.

Vision-based 3D semantic scene completion (SSC) attracted significant attention due to the increasing cost-effectiveness of purely visual autonomous driving solutions. MonoScene~\cite{cao2022monoscene} was the first to infer a scene's dense geometry and semantics from a single monocular RGB image.
TPVFormer~\cite{huang2023tri} proposed a three-perspective view representation that, together with BEV, provided two additional vertical planes. 
OccFormer~\cite{zhang2023occformer} was based on Lift-Splat-Shoot (LSS)~\cite{philion2020lift} used a dual-path transformer framework to encode 3D voxel features.
VoxFormer~\cite{li2023voxformer} utilized a novel two-stage framework to lift images into full 3D voxelized semantic scenes.
SurroundOcc~\cite{wei2023surroundocc} applied 3D convolutions to progressively upsample multi-scale voxel features and designed a pipeline to generate dense SSC ground truth.
BRGScene~\cite{li2023stereoscene} employed binocular image input to implicitly generate stereo depth information and applied stereo matching to resolve geometric ambiguities.
NDCScene~\cite{yao2023ndc} devised a novel Normalized Device Coordinates scene completion network that directly extended the 2D feature map into a Normalized Device Coordinates space rather than into world space.
Recently, MonoOcc~\cite{zheng2024monoocc} further enhanced the 3D volume with an image-conditioned cross-attention module.
H2GFormer~\cite{wang2024h2gformer} effectively utilized 2D features through a progressive feature reconstruction process across various directions.
Symphonize~\cite{jiang2024symphonize} extracted high-level instance features from the image feature map, serving as the key and value for cross-attention.
HASSC~\cite{wang2024HASSC} introduced a self-distillation training strategy to improve the performance of VoxFormer.
BRGScene~\cite{li2023stereoscene} utilized binocular image inputs to implicitly generate stereo depth information and employed stereo matching to resolve geometric ambiguities.
\subsection{Depth Estimation} 
Depth estimation methods were categorized based on the number of images utilized: single-view depth estimation and multi-view depth estimation. Predicting depth from a single image was inherently challenging, but leveraging contextual signals facilitated object depth estimation to some extent.
AdaBins~\cite{bhat2021adabins} utilized a CNN approach to predict depth from a single image. 
For multi-view depth estimation, constructing temporal stereo was an effective method for depth prediction.
MVSNet~\cite{yao2018mvsnet} first proposed constructing a differentiable cost volume and then used 3D CNNs to regularize the cost volume, achieving state-of-the-art accuracy at the time.
Fast-MVSNet~\cite{Yu_2020_fastmvsnet} employed sparse temporal stereo and Gaussian-Newton layers to enhance the speed of MVSNet.
Recently, BEVStereo~\cite{li2023bevstereo} introduced a dynamic temporal stereo technique to dynamically select the scale of matching candidates.
Li \textit{et al.}~\cite{Li_2023_CVPR} proposed a novel method to fuse multi-view and monocular cues encoded as volumes without relying on heuristically crafted masks.

\subsection{Large MultiModal Models}
The field of Large Multimodal Models (LMMs) experienced significant advancements, particularly in enhancing visual and language processing. Methods such as Flamingo~\cite{alayrac2022flamingo} advanced visual representation by integrating a Perceiver Resampler with vision encoders. BLIP2~\cite{li2023blip} employed a Q-Former to connect the frozen LLM and vision encoder. DriveCLIP~\cite{hasan2024vision} proposed a CLIP-based driver activity recognition approach that identifies driver distraction from naturalistic driving images and videos.
MiniGPT4\cite{zhu2023minigpt} bridged visual modules and LLMs, thereby enhancing multimodal capabilities. InstructBLIP~\cite{NEURIPS2023InstructBLIP}, building on BLIP2, introduced instructional inputs to the Q-Former for extracting task-relevant visual features.
Contrastive Vision-Language Pre-training (CLIP)\cite{radford2021clip} achieved significant progress in open-set and zero-shot image classification tasks due to the strong alignment between visual and text embeddings. LSeg\cite{li2022lseg} focused on aligning pixel-level image features with class embeddings generated by the CLIP text encoder, while OpenSeg~\cite{ghiasi2022openseg} aligned segment-level image features with image captions using a region-word basis. MaskCLIP~\cite{dong2023maskclip} explored dense prediction problems using CLIP by making minor adjustments to the network structure.
However, practical applications such as autonomous driving and indoor navigation required a deeper understanding of 3D scenes. Consequently, recent research explored the application of 2D vision-language pre-training for 3D perception tasks. For example, CLIP2Scene~\cite{chen2023clip2scene} introduced a semantically driven cross-modal contrastive learning framework, while OpenScene~\cite{Peng2023OpenScene} utilized a pre-trained VL model~\cite{ghiasi2022openseg, kuo2023F-VLM} to extract per-pixel CLIP features and project 3D points onto the image plane to derive dense 3D features.

\section{Methodology}
\subsection{Preliminary}
\paragraph{Problem setup}
Given a set of RGB images $I_t = \{I_{t}^{l}, I_{t}^{r}, I_{t-1}^{l}, I_{t-2}^{l}, ..., I_{t-i}^{l}\}$, where $I_t^l$ represents the reference image, $I_t^r$ represents the corresponding right eye image, and other are n-1 source images $\{I_{t-i}^l\}_{i=1}^{n-1}$. The goal is to infer the geometry and semantics of the 3D scene jointly. The scene is represented as a voxel grid ${Y}\in\mathbb{R}^{X \times Y \times Z \times{(M+1)} }$, where X, Y, Z represent height, width and depth in 3D space. For each voxel, it will be assigned to a unique semantic label belonging to $C\in\{C_0, C_1, ..., C_M\}$ that either occupies the empty space $C_0$ or falls in a specific semantic class $\{C_1, ..., C_M\}$. Here M represents the total number of semantic classes. We want to learn a transformation $Y=\theta(I_t)$ as close to the ground truth $\hat{Y}$ as possible.
\begin{figure*}[t]
\centering
  \includegraphics[width=\textwidth]{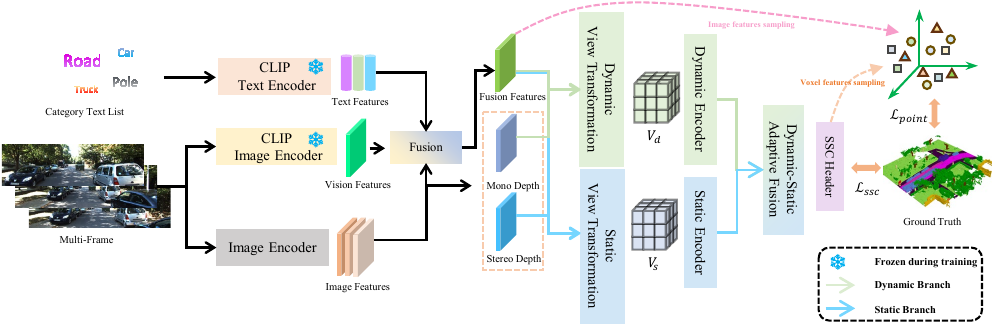}
  \caption{The CDScene framework is proposed for camera-based 3D semantic scene completion. The pipeline includes several key components. 
  }
  \vspace{-2mm}
  \label{fig:fig2}
\end{figure*}
\paragraph{Revisiting CLIP}
CLIP alleviates the shortcomings of expensive training data labeling and weak model generalization in the field of computer vision. CLIP consists of an image encoder (ResNet~\cite{he2016resnet} or ViT~\cite{dosovitskiy2020vit}) and a text encoder (Transformer~\cite{vaswani2017attention}), both respectively project the image and text representation to a joint embedding space. CLIP can achieve promising open-vocabulary recognition. For 2D zero-shot classification, CLIP first places the class name into a pre-defined template to generate the text embeddings and then encodes images to obtain image embeddings. Next, it calculates the similarities between image and text embeddings to determine the class.
LSeg~\cite{li2022lseg} uses a text encoder to compute embeddings of descriptive input labels together with a transformer-based image encoder that computes dense per-pixel embeddings of the input image. The image encoder is trained with a contrastive objective to align pixel embeddings to the text embedding of the corresponding semantic class. The text embeddings provide a flexible label representation in which semantically similar labels map to similar regions in the embedding space.
\paragraph{Overview}
We illustrate our method in Fig.~\ref{fig:fig2}. First, we use the lightweight image encoder as the backbone of the RGB image and obtain the image features $F_t\in\mathbb{R}^{C\times H\times W}$. The reference image features are passed through a deep network to obtain the monocular discrete depth distribution $D_{mono}$. The reference and source image features are transformed into a stereo depth distribution $D_{stereo}$ through the homography warping operation.
Meanwhile, we use CLIP to extract the visual features $F_{vis}$ and text features $text$ (sec~\ref{sec:lmms}). They are fused with the image features $F_t^l$ to obtain $F_{fusion}$.
Then, $D_{mono}$ and $F_{fusion}$ enter the dynamic branch (sec~\ref{method:Dynamic}) to obtain the dynamic voxel features $V_{d}$. $D_{stereo}$ and $F_{fusion}$ enter the static branch (sec~\ref{method:Static}) to obtain the static voxel features $V_{s}$. Then $V_{d}$ and $V_{s}$ enter two encoders composed of sparse convolutions respectively. The features of the two branches are adaptively fused (sec~\ref{method:cross}) to obtain the fused voxel feature $V_{fusion}$. Finally, $V_{fusion}$ outputs dense semantic voxels $Y$ through upsampling and linear projection.

\subsection{LMMs Features Extraction}
\label{sec:lmms}
\begin{figure}[t]
  \centering
  \includegraphics[width=0.75\linewidth]{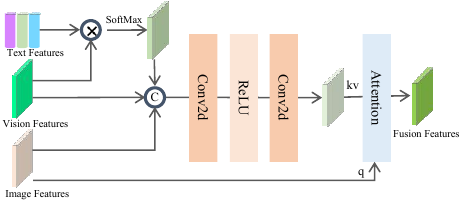}
  \caption{Illustration of the feature fusion module.}
  \vspace{-2mm}
  \label{fig:fig2_3}
\end{figure}

Given the input image, we extract image features $F_t^l$ and visual features $F_{vis}$ from standard image encoders~\cite{wang2023repvit} and CLIP image encoder~\cite{li2022lseg}.
Simultaneously, we utilize the category names as prompts to the text encoder, yielding the text features $text \in \mathbb{R}^{Q\times C}$, where $Q$ represents the number of categories and $C$ denotes the feature channels. Image features $F_t^l$ are further encoded into context features $F_{con}$ by the convolutional network.
{Regarding the CLIP visual features $F_{vis}$, we utilize the pre-trained clip\_vitl16\_384 variant and extract the output from the final transformer layer of the image encoder. These features are then projected to a $C$-dimensional space to ensure consistency with the dimensionality of the text embeddings, thereby enabling effective feature fusion and alignment in the downstream modules.
As shown in Fig.~{\ref{fig:fig2_3}}, we describe in detail the fusion process of text, visual and image features.}
First, we compute the cosine similarity between the CLIP image and text features to generate a 2D semantic map $F_{sem}$.
{By incorporating text embeddings alongside image features, we enable the model to more effectively align and match visual content with high-level semantic concepts, particularly in cases where visual cues alone may be ambiguous or insufficient—such as occluded or dynamically changing regions. In this way, the text features serve as semantic anchors, enhancing the model's understanding of the visual scene in a class-aware manner.}

{Then, we concatenate the three features $F_{sem}$, $F_{vis}$, and $F_{con}$ along the channel dimension to obtain a joint multi-modal feature representation. This concatenated feature is denoted as:}
\begin{equation}
   {F_{cat} = \operatorname{Concat}(F_{sem},F_{vis},F_{con}),}
\end{equation}
{To unify the feature spaces and enable effective interaction among these modalities, we perform an initial fusion by applying two sequential 2D convolutional layers with ReLU activation:}
\begin{equation}
    {\hat{F}_{vis} = {Conv}(ReLU(Conv(F_{cat}))),}
\end{equation}
{This operation projects the concatenated feature into a shared latent space, filters redundant information, and learns a preliminary fused representation $\hat{F}_{vis}$ that integrates semantic, visual, and contextual cues.
To further enhance the semantic alignment of visual features, we apply a cross-attention mechanism,}
\begin{equation}
   { F_{fusion} = \operatorname{Cross-Att}(Q = F_{con}, KV=\hat{F}_{vis}).}
\end{equation}
{This cross-attention allows the model to selectively emphasize semantically relevant visual features. The final output $F_{fusion}$ carries enriched semantic representation and serves as input for subsequent 3D scene reasoning and completion modules.}

To obtain point-wise image features, we sample them into point space,
\begin{equation}
    F_{point} = \operatorname{GridSample}(F_{fusion},P_{u,v}),
\end{equation}
where ${P}_{u,v}$ is the reference point coordinate.
\subsection{Dynamic Branch}
\label{method:Dynamic}
The dynamic branch mainly analyzes the information present in mono-depth features. 
{Monocular depth estimation captures rich appearance-based cues such as shading, texture gradients, object contours, and learned category priors. These cues are particularly valuable in scenarios where geometric disparity is unreliable or unavailable, such as in low-texture regions, occlusions, or rapidly moving objects, where stereo depth often fails due to lack of correspondence.
By incorporating monocular depth into the dynamic branch, we enable the model to capture shape, motion tendencies, and spatial relationships of dynamic entities through learned perspective and appearance priors. This strategy not only compensates for stereo estimation failures in non-rigid or occluded regions, but also allows the dynamic branch to operate effectively without requiring temporal alignment or stereo image pairs.}

Initially, we adopt the LSS paradigm~\cite{philion2020lift} for the 2D-3D view conversion process. Specifically, using the features ${F}_{t}^l$ of the left image, we derive a monocular discrete depth distribution $D_{mono}$. Subsequently, these outputs are further processed.
At the same time, following the method proposed in BRGScene~\cite{li2023stereoscene}, we use the binocular stereo constructor~\cite{guo2019gwcnet} to convert the features $F_t^l$ and $F_t^r$ of the left and right images into a dense 3D volume represented as $F_{bino}$. Then, this volume $F_{bino}$ interacts with the discrete depth distribution $D_{mono}$ to produce $\hat{D}_{mono}$. The specific calculation process is as follows:
\begin{equation}
   \hat{D}_{mono} = \psi(F_{bino},D_{mono}),
\end{equation}
where $\psi$ is the MIE module~\cite{li2023stereoscene}, which can selectively filter out the most reliable information from binocular and monocular depth features. Subsequently, we compute ${F}_{fusion} \otimes \hat{D}_{mono}$ to generate a point cloud representation denoted as $P \in\mathbb{R}^{N \times D \times h \times w \times C_{con}}$. Employing voxel pooling, We generate 3D feature volumes with dynamic cues ${V}_{d}$ The specific calculation process is as follows:
\begin{equation}
   {V}_{d} = \operatorname{VoxelPooling}({F}_{fusion},\hat{D}_{mono}).
\end{equation}
Next, the voxel features are guided to the dynamic encoder module to refine the dynamic voxel features. Subsequently, we use the dynamic-static adaptive fusion module to effectively aggregate the static voxel features with the dynamic voxel features.

\subsection{Static Branch}
\label{method:Static}
The static branch processes stereo depth information from consecutive image frames to understand the motion and evolution of the scene. 
{Stereo depth estimation, by exploiting disparity cues between two or more aligned temporal views, enables the precise extraction of dense and accurate 3D geometric information. This is especially crucial for capturing the fine-grained, spatial details of static objects and surfaces—such as roads, buildings, and other immobile elements—where texture and geometric structure are consistent across consecutive frames. Unlike monocular depth, which may struggle with precise geometric fidelity, stereo depth offers a high-resolution, dense depth map that is critical for accurate 3D scene reconstruction in well-textured regions.
Furthermore, stereo depth provides reliable spatial cues in areas where monocular depth can fail due to lack of sufficient texture or in occluded regions. The disparity between two images captured at slightly different viewpoints is particularly effective at revealing depth discontinuities and the fine geometric details of static structures. This information is indispensable for constructing dense static geometry and ensuring accurate representation of static parts of the scene, which are integral to the overall 3D scene completion process.
In the static branch, we combine stereo depth maps with homography deformation operations to obtain the features from consecutive image frames. This allows us to align and register the reference and deformed source features, which are then used to build a stereo cost volume. The resulting stereo depth map ensures precise depth inference and captures the necessary 3D structure of the static scene.}

Specifically, the reference image features $F_t^l$ and source image features $\{F_{t-i}^l\}_{i=1}^{n-1}$ first follow the BEVStereo~\cite{li2023bevstereo} multi-frame stereo depth construction paradigm, predicting depth from the mono feature (mono depth) and predicting depth from temporal stereo features (stereo depth). For mono depth, the depth prediction is predicted directly. For stereo depth, the depth center and depth range is first predicted and used to generate the stereo depth distribution. Additionally, a weight network is used to create a weight map that will be applied to the stereo depth. Mono depth and weighted stereo depth are combined for the ultimate multi-view stereo depth $D_{stereo}$. Then, we follow the same process as the static branch, interact $D_{stereo}$ with $F_{bino}$ to obtain the $\hat{D}_{stereo}$, and perform voxel pooling to obtain the static voxel feature $V_{s}$. 
\begin{figure}[t]
  \centering
  \includegraphics[width=0.75\linewidth]{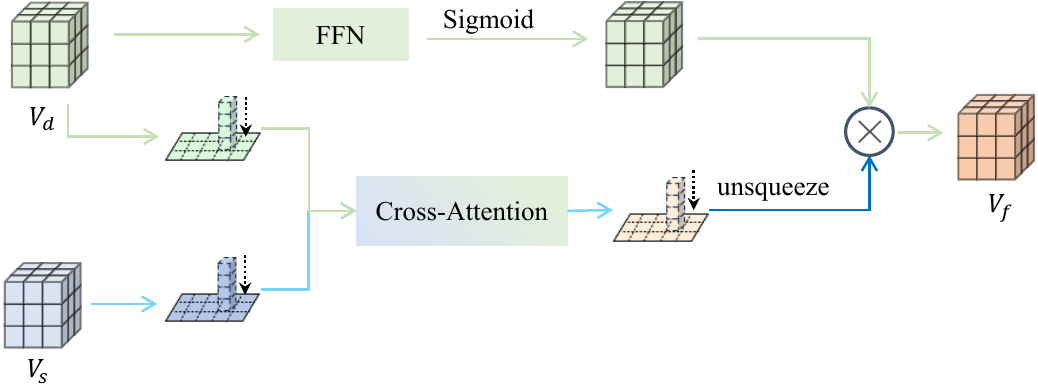}
  \caption{Illustration of the dynamic-static adaptive fusion module.}
  \vspace{-2mm}
  \label{fig:fig4}
\end{figure}

\subsection{Dynamic-Static Adaptive Fusion}
\label{method:cross}
{Dynamic voxel features obtained from the dynamic branch carry critical structural information about the motion and temporal evolution of dynamic objects. These features capture the dynamic regional structure, including the shape and motion patterns of dynamic elements in the scene. 
Static voxel features, obtained from the static branch, contain rich information about the geometric structure and spatiotemporal semantic relationships of static objects and surfaces. These features are instrumental in capturing the dense static geometry and textural details of the scene, which are fundamental for accurate reconstruction of background environments. }Therefore, we design the dynamic-static adaptive fusion module, as shown in Fig.~\ref{fig:fig4}. {By combining dynamic and static features, the fusion process establishes a spatiotemporal consistency in the scene. The dynamic features ensure that moving objects are accurately captured, while the static features maintain the integrity of the environment's geometry. }
To better express position and dynamic cues, through averaging pooling along the height dimension $z$, we derive dynamic BEV features ${F}_{d}\in\mathbb{R}^{C \times x \times y}$ and static BEV features ${F}_{s}\in\mathbb{R}^{C \times x \times y}$, corresponding to the two branches. Initially, self-attention is applied to ${F}_{d}$ to grasp the semantic structure of dynamic scenes. Subsequently, the refined $\hat{{F}}_{d}$ functions as a query, with the key and value derived from ${F}_{s}$, facilitating cross-attention. Given the necessity for relevant features to align across both branches, we prioritize local information near specific queries to establish cross-attention within the BEV context. Following this, the fusion of BEV and voxel characteristics occurs through a sigmoid-weighted sum residual connection, yielding fused voxel features.


Specifically, we first encode voxel features using convolutional layers, followed by average pooling along the height dimension. And represent it as a sequence of m-dimensional feature vectors, for example the feature sequence from the static branch is $F_{d}\in\mathbb{R}^{n_q\times C}$, where $n_q = x \times y$. 
First, self-attention~\cite{vaswani2017attention} is performed on the tokenized features $F_{d}$ to capture the dynamic scene semantic layout feature. $F_{d}$ are linearly projected to queries, keys and values: $\{Q_{d}, K_{d}, V_{d}\} \in\mathbb{R}^{n_q\times C}$ respectively. The calculation of the self-attention SA module is as follows:
\begin{equation}
  SA(F_{d}) = \operatorname{softmax}(\frac{{Q_d} \cdot{K_d}}{\sqrt{d}}) \cdot V_d,
\end{equation}
where $d$ is the scaling factor, the refined features $\hat{{F}}_{d}$ are obtained after the self-attention operation.
We perform neighborhood cross-attention on refined dynamic features $\hat{{F}}_{d}$, inspired by neighborhood attention~\cite{Hassani_2023_Neighborhood}.
Subsequently, the refined features $\hat{{F}}_{d}$ are linearly projected to the query $\hat{Q_{d}}\in\mathbb{R}^{n_q\times C}$.
For a sequence of tokenized features in the multi-frame branch, use a linear layer to project it to keys $K_{s}\in\mathbb{R}^{n_q\times C}$ and values $V_{s} \in\mathbb {R}^{n_q\times C}$. The cross attention ${CA}^i$ of query $i$ in the dynamic feature map is calculated as follows:
\begin{equation}
    {CA}^i=\operatorname{softmax}\left(\frac{Q_d^i \cdot\left(K_s^{\rho(i)}\right)^T+B^{(i, \rho(i))}}{\sqrt{d}}\right) \cdot V_s^i,
\end{equation}
where $\rho(i)$ is the neighborhood with the size of $k$ centered at the same position in the multi-frame branch, $B^{(i, \rho(i))}$ denotes the positional bias added to the attention and $d$ is the scaling factor. When multi-head attention is applied, the outputs of each head are concatenated. For each pixel in the feature map, we calculate the cross attention as above. The final feature maps ${F}_{f}$ become dense and rich in information.
Then, we apply ASPP~\cite{chen2017aspp} to capture the global context. Finally, bev-level information is propagated throughout the 3D voxels. The output fused voxel feature ${V}_{f}$ is computed as:
\begin{equation}
    {V}_{f} = {V}_{d} + \text{ASPP}({F}_{f}) \cdot \delta(\text{FFN}({V}_{d})),
\end{equation}
where $\delta(\cdot)$ is the sigmoid function, FFN for generating the aggregation weight along the height dimension.
To obtain point-wise voxel features, we sample them into point space,
\begin{equation}
    V_{point} = \operatorname{GridSample}(V_{f},\mathbf{P}_{sample}),
\end{equation}
where $\mathbf{P}_{sample}$ is the reference point coordinate.
Finally, the point-wise loss function is as follows:
\begin{equation}
    \begin{split}
        \mathcal{L}_{point} = \mathcal{L}_{ce}(\hat{Y},\text{MLP}(\text{concat}(F_{point},V_{point})))+\\
        \mathcal{L}_{lovasz}(\hat{Y},\text{MLP}(\text{concat}(F_{point},V_{point}))).
    \end{split}
\end{equation}

\subsection{Training Loss}
In the CDScene framework, we adopt the Scene-Class Affinity Loss $\mathcal{L}_{scal}$ from MonoScene~\cite{cao2022monoscene} to optimize precision, recall, and specificity concurrently. Besides, the intermediate depth distribution for view transformation is supervised by the projections of LiDAR points, with the binary cross-entropy loss $\mathcal{L}_{depth}$ following BEVDepth~\cite{li2023bevdepth}.
The SSC loss function is formulated as follows: 
\begin{equation}
    \mathcal{L}_{ssc} =\mathcal{L}^{sem}_{scal} + \mathcal{L}^{geo}_{scal} + \mathcal{L}_{ce} +  \mathcal{L}_{depth}^{d} + \mathcal{L}_{depth}^{s}.
\end{equation}
The overall loss function is as follows:
\begin{equation}
    \mathcal{L} = \lambda_{ssc}\mathcal{L}_{ssc}+ \lambda_{point}\mathcal{L}_{point},
\end{equation}
where several $\lambda_s$ are balancing coefficients.

\section{Experiments}
\label{sec:exp}
To assess the effectiveness of our CDScene, we conducted thorough experiments using the large outdoor datasets SemanticKITTI~\cite{behley2019semantickitti,Geiger2012kitti}, SSCBench-KITTI-360~\cite{li2023sscbench,Liao2022kitti360} and SemanticKITTI-C.
\subsection{Experimental Setup}
\label{detail}
\paragraph{Datasets} 
The SemanticKITTI~\cite{behley2019semantickitti,Geiger2012kitti} dataset has dense semantic scene completion annotations and labels a voxelized scene with 20 semantic classes. The dataset comprises 10 training sequences, 1 validation sequence, and 11 testing sequences. RGB images are resized to $1280 \times 384$ dimensions to serve as inputs.
The SSCBench-KITTI-360~\cite{li2023sscbench,Liao2022kitti360} dataset provides 7 training sequences, 1 validation sequence, and 1 testing sequence, covering a total of 19 semantic classes. The RGB images are resized to $1408 \times 384$ resolution for input processing. 
In addition, to verify the robustness of our method, we performed a degradation operation on the image data of the SemanticKITTI validation set~\cite{kong2023robodepth}, forming SemanticKITTI-C. This dataset contains 6 corruption types of weather and lighting conditions, with severity levels of 1, 3, and 5 for each category.

\paragraph{Metrics} Following \cite{song2017semantic}, the main consideration of SSC is the mean Intersection over Union (mIoU), which considers the IoU of all semantic classes for prediction without considering the free space.
The mIoU is calculated by:
\begin{equation}
  mIoU = \frac{1}{C} \sum^{C}_{c=1}\frac{TP_c}{TN_c+FP_c+FN_c}
\end{equation}
Here, $TP_c$, $TN_c$, $FP_c$, and $FN_c$ are the true positives, true negatives, false positives and false negatives predictions for class $c$.
The SSC task requires considering pure geometric reconstruction quality, while mIoU takes into account all semantic categories. Therefore, Intersection over Union (IoU), Precision, and Recall are often used to represent scene representations indicating empty or occupied areas.
\paragraph{Implementation Details} 
We use RepVit~\cite{wang2023repvit} and FPN~\cite{lin2017feature} to extract features for all images.
For dynamic view transformation, we use the LSS paradigm for 2D-3D projection. We use the BEVStereo paradigm for static view transformation. For dynamic-static adaptive fusion, a total of 3 self-attention layers and two cross-attention layers are included. The neighborhood attention range is set to 7, and the number of attention heads is set to 8. Finally, the final outputs of SemantiKITTI and SemantiKITTI-C are 20 classes, and SSCBench-KITTI-360 is 19 classes. All datasets have the scene size of $51.2m \times 51.2m \times 64m$ with the voxel grid size of $256 \times 256 \times 32$. By default, the model is trained for 30 epochs. We optimise the process, utilizing the AdamW optimizer with an initial learning rate of 1e-4 and a weight decay of 0.01. We also employ a multi-step scheduler to reduce the learning rate. All models are trained on two A100 Nvidia GPUs with 80G memory and batch size 4.



\begin{table*}[t]
  \centering
  \renewcommand
  \arraystretch{1.2}
  \setlength{\tabcolsep}{2pt}
    \caption{Quantitative results on the SemanticKITTI hidden test set. {\textbf{Bold}} denotes the best performance.}
  \resizebox{\textwidth}{!}{
  \begin{tabular}{l|l|c|rrrrrrrrrrrrrrrrrrr|r}
    \toprule
    \textbf{Methods} &\textbf{Venues}&\textbf{IoU}   
    & \rotatebox{90}{\vcenteredbox{\colorbox[RGB]{255,0,255}{\textcolor[RGB]{255,0,255}{\rule{1px}{1px}}}} \textbf{road} (15.30$\%$)} 
    & \rotatebox{90}{\vcenteredbox{\colorbox[RGB]{75,0,75}{\textcolor[RGB]{75,0,75}{\rule{1px}{1px}}}} \textbf{sidewalk} (11.13$\%$)} 
    & \rotatebox{90}{\vcenteredbox{\colorbox[RGB]{255,150,255}{\textcolor[RGB]{255,150,255}{\rule{1px}{1px}}}} \textbf{parking} (1.12$\%$)} 
    & \rotatebox{90}{\vcenteredbox{\colorbox[RGB]{175,0,75}{\textcolor[RGB]{175,0,75}{\rule{1px}{1px}}}} \textbf{other-grnd} (0.56$\%$)} 
    &  \rotatebox{90}{\vcenteredbox{\colorbox[RGB]{255,200,0}{\textcolor[RGB]{255,200,0}{\rule{1px}{1px}}}} \textbf{building} (14.10$\%$)} 
    &  \rotatebox{90}{\vcenteredbox{\colorbox[RGB]{100,150,245}{\textcolor[RGB]{100,150,245}{\rule{1px}{1px}}}} \textbf{car} (3.92$\%$)} 
    & \rotatebox{90}{\vcenteredbox{\colorbox[RGB]{80,30,180}{\textcolor[RGB]{80,30,180}{\rule{1px}{1px}}}} \textbf{truck} (0.16$\%$)} 
    & \rotatebox{90}{\vcenteredbox{\colorbox[RGB]{100,230,245}{\textcolor[RGB]{100,230,245}{\rule{1px}{1px}}}} \textbf{bicycle} (0.03$\%$)}  
    & \rotatebox{90}{\vcenteredbox{\colorbox[RGB]{30,60,150}{\textcolor[RGB]{30,60,150}{\rule{1px}{1px}}}} \textbf{motocycle} (0.03$\%$)} 
    & \rotatebox{90}{\vcenteredbox{\colorbox[RGB]{0,0,255}{\textcolor[RGB]{0,0,255}{\rule{1px}{1px}}}} \textbf{other-vehicle} (0.20$\%$)} 
    & \rotatebox{90}{\vcenteredbox{\colorbox[RGB]{0,175,0}{\textcolor[RGB]{0,175,0}{\rule{1px}{1px}}}} \textbf{vegetation} (39.3$\%$)} 
    & \rotatebox{90}{\vcenteredbox{\colorbox[RGB]{135,60,0}{\textcolor[RGB]{135,60,0}{\rule{1px}{1px}}}}  \textbf{trunk} (0.51$\%$)} 
    & \rotatebox{90}{\vcenteredbox{\colorbox[RGB]{150,240,80}{\textcolor[RGB]{150,240,80}{\rule{1px}{1px}}}} \textbf{terrain} (9.17$\%$)} 
    & \rotatebox{90}{\vcenteredbox{\colorbox[RGB]{255,30,30}{\textcolor[RGB]{255,30,30}{\rule{1px}{1px}}}} \textbf{person} (0.07$\%$)} 
    & \rotatebox{90}{\vcenteredbox{\colorbox[RGB]{255,40,200}{\textcolor[RGB]{255,40,200}{\rule{1px}{1px}}}} \textbf{bicylist} (0.07$\%$)} 
    & \rotatebox{90}{\vcenteredbox{\colorbox[RGB]{150,30,90}{\textcolor[RGB]{150,30,90}{\rule{1px}{1px}}}}  \textbf{motorcyclist} (0.05$\%$)} 
    &  \rotatebox{90}{\vcenteredbox{\colorbox[RGB]{255,120,50}{\textcolor[RGB]{255,120,50}{\rule{1px}{1px}}}} \textbf{fence} (3.90$\%$)} 
    & \rotatebox{90}{\vcenteredbox{\colorbox[RGB]{255,240,150}{\textcolor[RGB]{255,240,150}{\rule{1px}{1px}}}} \textbf{pole} (0.29$\%$)} 
    & \rotatebox{90}{\vcenteredbox{\colorbox[RGB]{255,0,0}{\textcolor[RGB]{255,0,0}{\rule{1px}{1px}}}} \textbf{traf.-sign} (0.08$\%$)}& \textbf{mIoU}  \\
    
    \midrule
    MonoScene~\cite{cao2022monoscene}&CVPR'2022 & 34.16  & 54.70&27.10&24.80&5.70&14.40&18.80&3.30&0.50&0.70&4.40&14.90&2.40&19.50&1.00&1.40&0.40&11.10&3.30&2.10 & 11.08 \\
    
    TPVFormer~\cite{huang2023tri}& CVPR'2023& 34.25  &55.10&27.20&27.40&6.50&14.80&19.20&3.70&1.00&0.50&2.30&13.90&2.60&20.40&1.10&2.40&0.30&11.00&2.90&1.50& 11.26\\
    SurroundOcc~\cite{wei2023surroundocc}& ICCV'2023&34.72&56.90&28.30&30.20&6.80&15.20&20.60&1.40&1.60&1.20&4.40&14.90&3.40&19.30&1.40&2.00&0.10&11.30&3.90&2.40&11.86\\

    OccFormer~\cite{zhang2023occformer}& ICCV'2023 & 34.53  & 55.90&30.30&31.50&6.50&15.70&21.60&1.20&1.50&1.70&3.20&16.80&3.90&21.30&2.20&1.10&0.20&11.90&3.80&3.70& 12.32\\
    VoxFormer-T~\cite{li2023voxformer}& CVPR'2023& 43.21& 54.10& 26.90& 25.10& 7.30& 23.50& 21.70& 3.60& 1.90& 1.60& 4.10& 24.40& 8.10& 24.20& 1.60& 1.10& 0.00& 13.10& 6.60& 5.70& 13.41\\
    MonoOcc~\cite{zheng2024monoocc}&ICRA'2024&-&55.20&27.80&25.10&9.70&21.40&23.20&\textbf{5.20}&2.20&1.50&5.40&24.00&8.70&23.00&1.70&2.00&0.20&13.40&5.80&6.40&13.80\\
    H2GFormer~\cite{wang2024h2gformer}&AAAI'2024&44.20&56.40&28.60&26.50&4.90&22.80&23.40&4.80&0.80&0.90&4.10&24.60&9.10&23.80&1.20&2.50&0.10&13.30&6.40&6.30&13.72\\
    HASSC~\cite{wang2024HASSC}&CVPR'2024&43.40&54.60&27.70&23.80&6.20&21.10&22.80&4.70&1.60&1.00&3.90&23.80&8.50&23.30&1.60&4.00&0.30&13.10&5.80&5.50&13.34\\
    Symphonize~\cite{jiang2024symphonize}&CVPR'2024&42.19&58.40&29.30&26.90&\textbf{11.70}&24.70&23.60&3.20&\textbf{3.60}&2.60&5.60&24.20&10.00&23.10&\textbf{3.20}&1.90&\textbf{2.00}&16.10&7.70&8.00&15.04\\
    BRGScene~\cite{li2023stereoscene}&IJCAI'2024 & 43.34 & 61.90&31.20  &30.70 & 10.70 & 24.20 & 22.80 & 2.80 & 3.40 & 2.40 & 6.10 & 23.80 & 8.40 & 27.00 & 2.90 & 2.20 & 0.50 & 16.50 & 7.00 & 7.20 & 15.36\\
    \hline
    \rowcolor{gray!20}\textbf{Ours}&  & \textbf{45.60} &\textbf{62.56}&\textbf{34.69}&\textbf{31.80}&11.51&\textbf{28.04}&\textbf{25.85}&5.11&3.01&\textbf{2.92}&\textbf{7.75}&\textbf{26.47}&\textbf{10.62}&\textbf{30.21}&2.89&\textbf{5.61}&1.30&\textbf{19.15}&\textbf{9.55}&\textbf{8.62}& \textbf{17.24}\\

    \bottomrule
  \end{tabular}}

  \label{tab:Test Quantitative Comparison}
\end{table*}
\begin{table*}[t]
  \centering
  \renewcommand\arraystretch{1.2}
  \setlength{\tabcolsep}{2pt}
    \caption{Quantitative results on the SSCBench-KITTI360 test set. {\textbf{Bold}} denotes the best performance.}
  \resizebox{\textwidth}{!}{
  \begin{tabular}{l|rrr|rrrrrrrrrrrrrrrrrr|r}
    \toprule
    \textbf{Methods}&\textbf{Prec.}&\textbf{Rec.} &\textbf{IoU}   
    &  \rotatebox{90}{\vcenteredbox{\colorbox[RGB]{100,150,245}{\textcolor[RGB]{100,150,245}{\rule{1px}{1px}}}} \textbf{car} (2.85$\%$)} 
    & \rotatebox{90}{\vcenteredbox{\colorbox[RGB]{100,230,245}{\textcolor[RGB]{100,230,245}{\rule{1px}{1px}}}} \textbf{bicycle} (0.01$\%$)}  
    & \rotatebox{90}{\vcenteredbox{\colorbox[RGB]{30,60,150}{\textcolor[RGB]{30,60,150}{\rule{1px}{1px}}}} \textbf{motocycle} (0.01$\%$)} 
    & \rotatebox{90}{\vcenteredbox{\colorbox[RGB]{80,30,180}{\textcolor[RGB]{80,30,180}{\rule{1px}{1px}}}} \textbf{truck} (0.16$\%$)} 
    & \rotatebox{90}{\vcenteredbox{\colorbox[RGB]{0,0,255}{\textcolor[RGB]{0,0,255}{\rule{1px}{1px}}}} \textbf{other-vehicle} (5.75$\%$)} 
    & \rotatebox{90}{\vcenteredbox{\colorbox[RGB]{255,30,30}{\textcolor[RGB]{255,30,30}{\rule{1px}{1px}}}} \textbf{person} (0.02$\%$)} 
    &  \rotatebox{90}{\vcenteredbox{\colorbox[RGB]{255,0,255}{\textcolor[RGB]{255,0,255}{\rule{1px}{1px}}}} \textbf{road} (14.98$\%$)} 
    & \rotatebox{90}{\vcenteredbox{\colorbox[RGB]{255,150,255}{\textcolor[RGB]{255,150,255}{\rule{1px}{1px}}}} \textbf{parking} (2.31$\%$)} 
    & \rotatebox{90}{\vcenteredbox{\colorbox[RGB]{75,0,75}{\textcolor[RGB]{75,0,75}{\rule{1px}{1px}}}} \textbf{sidewalk} (6.43$\%$)} 
    & \rotatebox{90}{\vcenteredbox{\colorbox[RGB]{175,0,75}{\textcolor[RGB]{175,0,75}{\rule{1px}{1px}}}} \textbf{other-grnd} (2.05$\%$)} 
    &  \rotatebox{90}{\vcenteredbox{\colorbox[RGB]{255,200,0}{\textcolor[RGB]{255,200,0}{\rule{1px}{1px}}}} \textbf{building} (15.67$\%$)} 
    &  \rotatebox{90}{\vcenteredbox{\colorbox[RGB]{255,120,50}{\textcolor[RGB]{255,120,50}{\rule{1px}{1px}}}} \textbf{fence} (0,96$\%$)} 
    & \rotatebox{90}{\vcenteredbox{\colorbox[RGB]{0,175,0}{\textcolor[RGB]{0,175,0}{\rule{1px}{1px}}}} \textbf{vegetation} (41.99$\%$)} 
    & \rotatebox{90}{\vcenteredbox{\colorbox[RGB]{150,240,80}{\textcolor[RGB]{150,240,80}{\rule{1px}{1px}}}} \textbf{terrain} (7.10$\%$)} 
    & \rotatebox{90}{\vcenteredbox{\colorbox[RGB]{255,240,150}{\textcolor[RGB]{255,240,150}{\rule{1px}{1px}}}} \textbf{pole} (0.22$\%$)} 
    & \rotatebox{90}{\vcenteredbox{\colorbox[RGB]{255,0,0}{\textcolor[RGB]{255,0,0}{\rule{1px}{1px}}}} \textbf{traf.-sign} (0.06$\%$)} 
    & \rotatebox{90}{\vcenteredbox{\colorbox[RGB]{0, 150, 255}{\textcolor[RGB]{0, 150, 255}{\rule{1px}{1px}}}}  \textbf{other-struct.} (4.33$\%$)} 
    & \rotatebox{90}{\vcenteredbox{\colorbox[RGB]{255, 255, 50}{\textcolor[RGB]{255, 255, 50}{\rule{1px}{1px}}}} \textbf{other-obj.} (0.28$\%$)} &\textbf{mIoU}   \\
    
    \midrule
    MonoScene&56.73 &53.26&37.87&19.34&0.43&0.58&8.02&2.03&0.86&48.35&11.38&28.13&3.32&32.89&3.53&26.15&16.75&6.92&5.67&4.20&3.09&12.31\\
    VoxFormer&58.52 &53.44&38.76&17.84&1.16&0.89&4.56&2.06&1.63&47.01&9.67&27.21&2.89&31.18&4.97&28.99&14.69&6.51&6.92&3.79&2.43&11.91\\
    TPVFormer&59.32 &55.54&40.22&21.56&1.09&1.37&8.06&2.57&2.38&52.99&11.99&31.07& 3.78&34.83&4.80&30.08&17.52&7.46&5.86&5.48&2.70&13.64\\
    OccFormer&59.70 &55.31&40.27&22.58&0.66&0.26&9.89&3.82&2.77&54.30&13.44&31.53&3.55&36.42&4.80&31.00&19.51&7.77&8.51& 6.95&4.60&13.81\\
    BRGScene&69.02& 57.53& 45.73 & 28.60 & 2.55 & 3.44 & 12.85 & 6.18 & 5.91 & 59.15 & 15.10 & 36.96 & 5.26 & 40.92 & 8.68 & 36.79 & 23.32 & 15.71 & 16.89 & 9.79 & 5.65 & 17.56\\
    Symphonies&69.24&54.88&44.12&\textbf{30.02}&1.85&\textbf{5.90}&\textbf{25.07}&\textbf{12.06}&\textbf{8.20}&54.94&13.83&32.76&\textbf{6.93}&35.11&8.58&\textbf{38.33}&11.52&14.01&9.57&\textbf{14.44}&\textbf{11.28}&18.58\\
    \hline
    \rowcolor{gray!20}\textbf{Ours}& \textbf{70.28} &\textbf{58.46} & \textbf{46.88} &29.51&\textbf{4.09}&5.12&18.53&6.50&7.41&\textbf{61.04}&\textbf{16.80}&\textbf{39.14}&6.14&\textbf{43.26}&\textbf{9.53}&37.11&\textbf{24.04}&\textbf{17.55}&\textbf{18.30}&10.52&7.09& \textbf{19.04}\\
    \bottomrule
  \end{tabular}}

  \label{tab:KITTI360Test Quantitative Comparison}
\end{table*}

\begin{table*}[t]
  \caption{Quantitative results on the SemanticKITTI validation set. The best results are in {\textbf{Bold}}.}
  \label{tab:kittival}
  \setlength{\tabcolsep}{2pt}
  \renewcommand\arraystretch{1.2}
  
  \resizebox{\textwidth}{!}{
  \begin{tabular}{l|l|r|rrrrrrrrrrrrrrrrrrr|r}
    \toprule
    \textbf{Methods} &\textbf{Published}&\textbf{IoU}   
    & \rotatebox{90}{\vcenteredbox{\colorbox[RGB]{255,0,255}{\textcolor[RGB]{255,0,255}{\rule{1px}{1px}}}} \textbf{road} (15.30$\%$)} 
    & \rotatebox{90}{\vcenteredbox{\colorbox[RGB]{75,0,75}{\textcolor[RGB]{75,0,75}{\rule{1px}{1px}}}} \textbf{sidewalk} (11.13$\%$)} 
    & \rotatebox{90}{\vcenteredbox{\colorbox[RGB]{255,150,255}{\textcolor[RGB]{255,150,255}{\rule{1px}{1px}}}} \textbf{parking} (1.12$\%$)} 
    & \rotatebox{90}{\vcenteredbox{\colorbox[RGB]{175,0,75}{\textcolor[RGB]{175,0,75}{\rule{1px}{1px}}}} \textbf{other-grnd} (0.56$\%$)} 
    &  \rotatebox{90}{\vcenteredbox{\colorbox[RGB]{255,200,0}{\textcolor[RGB]{255,200,0}{\rule{1px}{1px}}}} \textbf{building} (14.10$\%$)} 
    &  \rotatebox{90}{\vcenteredbox{\colorbox[RGB]{100,150,245}{\textcolor[RGB]{100,150,245}{\rule{1px}{1px}}}} \textbf{car} (3.92$\%$)} 
    & \rotatebox{90}{\vcenteredbox{\colorbox[RGB]{80,30,180}{\textcolor[RGB]{80,30,180}{\rule{1px}{1px}}}} \textbf{truck} (0.16$\%$)} 
    & \rotatebox{90}{\vcenteredbox{\colorbox[RGB]{100,230,245}{\textcolor[RGB]{100,230,245}{\rule{1px}{1px}}}} \textbf{bicycle} (0.03$\%$)}  
    & \rotatebox{90}{\vcenteredbox{\colorbox[RGB]{30,60,150}{\textcolor[RGB]{30,60,150}{\rule{1px}{1px}}}} \textbf{motocycle} (0.03$\%$)} 
    & \rotatebox{90}{\vcenteredbox{\colorbox[RGB]{0,0,255}{\textcolor[RGB]{0,0,255}{\rule{1px}{1px}}}} \textbf{other-vehicle} (0.20$\%$)} 
    & \rotatebox{90}{\vcenteredbox{\colorbox[RGB]{0,175,0}{\textcolor[RGB]{0,175,0}{\rule{1px}{1px}}}} \textbf{vegetation} (39.3$\%$)} 
    & \rotatebox{90}{\vcenteredbox{\colorbox[RGB]{135,60,0}{\textcolor[RGB]{135,60,0}{\rule{1px}{1px}}}}  \textbf{trunk} (0.51$\%$)} 
    & \rotatebox{90}{\vcenteredbox{\colorbox[RGB]{150,240,80}{\textcolor[RGB]{150,240,80}{\rule{1px}{1px}}}} \textbf{terrain} (9.17$\%$)} 
    & \rotatebox{90}{\vcenteredbox{\colorbox[RGB]{255,30,30}{\textcolor[RGB]{255,30,30}{\rule{1px}{1px}}}} \textbf{person} (0.07$\%$)} 
    & \rotatebox{90}{\vcenteredbox{\colorbox[RGB]{255,40,200}{\textcolor[RGB]{255,40,200}{\rule{1px}{1px}}}} \textbf{bicylist} (0.07$\%$)} 
    & \rotatebox{90}{\vcenteredbox{\colorbox[RGB]{150,30,90}{\textcolor[RGB]{150,30,90}{\rule{1px}{1px}}}}  \textbf{motorcyclist} (0.05$\%$)} 
    &  \rotatebox{90}{\vcenteredbox{\colorbox[RGB]{255,120,50}{\textcolor[RGB]{255,120,50}{\rule{1px}{1px}}}} \textbf{fence} (3.90$\%$)} 
    & \rotatebox{90}{\vcenteredbox{\colorbox[RGB]{255,240,150}{\textcolor[RGB]{255,240,150}{\rule{1px}{1px}}}} \textbf{pole} (0.29$\%$)} 
    & \rotatebox{90}{\vcenteredbox{\colorbox[RGB]{255,0,0}{\textcolor[RGB]{255,0,0}{\rule{1px}{1px}}}} \textbf{traf.-sign} (0.08$\%$)}& \textbf{mIoU}  \\
    
    \midrule
    MonoScene \cite{cao2022monoscene}&CVPR'2022 & 36.86&56.52&26.72&14.27&0.46&14.09&23.26&6.98&0.61&0.45&1.48&17.89&2.81&29.64&1.86&1.20&0.00&5.84&4.14&2.25&11.08 \\
    
    TPVFormer \cite{huang2023tri}& CVPR'2023& 35.61&56.50& 25.87& 20.60& 0.85& 13.88& 23.81& 8.08& 0.36& 0.05 &4.35 &16.92& 2.26& 30.38& 0.51& 0.89& 0.00& 5.94& 3.14& 1.52&11.36 \\
    
    OccFormer\cite{zhang2023occformer}& ICCV'2023 & 36.50  & 58.85& 26.88& 19.61& 0.31& 14.40& 25.09& 25.53& 0.81& 1.19& 8.52& 19.63& 3.93& 32.62& 2.78& 2.82& 0.00& 5.61& 4.26& 2.86 & 13.46\\
    VoxFormer-T\cite{li2023voxformer}& CVPR'2023& 44.15 &53.57&26.52&19.69&0.42&19.54&26.54&7.26&1.28&0.56&7.81&26.10&6.10&33.06&1.93&1.97&0.00&7.31&9.15&4.94 & 13.35\\
    HASSC~\cite{wang2024HASSC}&CVPR'2024&44.82& 57.05&28.25&15.90&1.04&19.05&27.23&9.91&0.92&0.86&5.61&25.48&6.15&32.94&2.80&\textbf{4.71}&0.00&6.58&7.68&4.05&13.48\\
    Symphonize~\cite{jiang2024symphonize}&CVPR'2024&41.92&56.37&27.58&15.28&0.95&21.64&28.68&20.44&2.54&2.82&\textbf{13.89}&25.72&6.60&30.87&3.52&2.24&0.00&8.40&9.57&5.76&14.89\\

    
    \midrule
    \rowcolor{gray!20}\textbf{Ours}&  & {\textbf{45.11}} &{\textbf{63.04}}&{\textbf{32.31}}& \textbf{24.37}&\textbf{2.67}& {\textbf{33.11}}& {\textbf{32.38}}&\textbf{26.07}&\textbf{2.94}& {\textbf{4.17}}&13.79& {\textbf{26.25}}& {\textbf{8.51}}& {\textbf{37.79}}&\textbf{4.17}& {{2.70}}& {\textbf{0.01}}& {\textbf{11.65}}& {\textbf{12.00}}& {\textbf{7.05}}&  {\textbf{17.83}}\\
    
    \bottomrule
  \end{tabular}}
\end{table*}
\begin{table*}
  \centering
  \caption{Robust Semantic Scene Completion results on the SemanticKITTI-C. {\textbf{Bold}} denotes the best performance.}
  \label{tab:robust result}
  \resizebox{0.75\textwidth}{!}{
  \begin{tabular}{cc|>{\columncolor{gray!15}}m{1cm}>{\columncolor{gray!15}}m{1cm}m{1cm}m{1cm}m{1cm}m{1cm}m{1cm}m{1cm}}
    \toprule
    \multirow{2}{*}{\textbf{Corruptions}} &\multirow{2}{*}{\textbf{Severity}}& \multicolumn{2}{c}{\textbf{Ours}} & \multicolumn{2}{c}{BRGScene} & \multicolumn{2}{c} {OccFormer} & \multicolumn{2}{c}{MonoScene}   \\
      & &{IoU} &{mIoU}&{IoU}&{mIoU}&{IoU}&{mIoU}&{IoU}&{mIoU}\\
    \midrule
     
     \multirow{3}{*}{\textbf{brightness}}&1 &\textbf{44.97} &	\textbf{16.40} &	42.70 &	14.93 &	36.30 &	13.45 & 36.83&	11.22\\
        &3 &\textbf{44.86} &	\textbf{15.66} 	&	42.05 &	14.17 	&	36.19 &	12.96 &		36.61&	10.77\\
        &5 &\textbf{43.78} &	\textbf{14.55} &	41.31 &	13.38 	&	35.96 &	12.02 		&36.16	&10.33 \\
\hline

     \multirow{3}{*}{\textbf{contrast}}&1 &\textbf{44.51} &	\textbf{14.4}8 	&	41.41 &	13.17 	&	35.77 &	11.88 	&	35.93& 	10.48 \\
        &3 &\textbf{43.32} &	\textbf{13.15} &		38.59 &	10.60 	&	34.44 &	10.09 	&	33.87 &	9.30 \\
        &5 &\textbf{40.72} &	\textbf{10.99} 	&	33.52 &	7.30 &		29.51 &	6.40 	&	28.88 &	6.10 \\       
\hline
        
     \multirow{3}{*}{\textbf{dark}}&1 &\textbf{41.42} &	\textbf{12.84} &		40.37 &	12.26 	&	34.81 &	10.86 	&	35.56 	&10.24 \\
        &3 &37.28 &	\textbf{9.85} 	&	\textbf{37.80} &	9.80 	&	32.70 &	8.50 	&	33.92 	&9.02\\
        &5 &29.81 &	5.10 	&	\textbf{32.51} &	5.57 	&	27.15 &	5.04 	&	29.01 	&\textbf{6.17}  \\   
\hline
        
     \multirow{3}{*}{\textbf{fog}}&1 &\textbf{44.01} &	\textbf{13.77} 	&	40.71 	&12.50 	&	35.60 &	11.59 	&	35.41 &	10.25 \\
        &3 &\textbf{43.24} &	\textbf{13.02} 	&	39.17 &	11.08 	&	34.90 &	10.70 	&	34.32 &	9.64 \\
        &5 &\textbf{41.70} &	\textbf{11.92} 	&	36.70 &	9.27 	&	34.07 &	9.92 	&	33.38 &	8.99  \\   
\hline
        
     \multirow{3}{*}{\textbf{frost}}&1 &\textbf{40.41} &	\textbf{12.21} 	&	37.56 &	10.44 	&	34.83 &	10.65 &		35.27& 	9.71 \\
        &3 &\textbf{34.42} &	\textbf{8.48} 	&	30.31 &	6.44 	&	31.34 &	7.45 &		31.90 &	7.61 \\
        &5 &\textbf{32.08} &	\textbf{7.30} 	&	27.82 &	5.51 	&	29.62 &	6.41 &		30.15& 	6.65  \\   
\hline
        
     \multirow{3}{*}{\textbf{snow}}&1 &\textbf{40.67} &	\textbf{13.35} 	&	38.52 &	12.13 	&	34.77 &	11.05 	&	34.94 &	9.76 \\
        &3 &\textbf{36.19} &	\textbf{10.65} 	&	34.81 &	9.65 	&	32.02 &	7.86 	&	32.80 &	8.34 \\
        &5 &\textbf{33.40} &	\textbf{8.63} 	&	32.46 &	7.47 	&	29.44 &	5.99 	&	30.41 &	6.69  \\   
    \hline
    \multicolumn{2}{c|}{\textbf{mean}}  & \textbf{39.82} &	\textbf{11.80} 	&	37.13 &	10.32 	&	33.30 	&9.60 	&	33.63 &	8.96 \\
    \bottomrule
  \end{tabular}}
\end{table*}
\begin{table}[ht]
  \centering
    \caption{Comparison of different ranges on SemanticKITTI val set.}
  \label{tab:diff range}

  \resizebox{0.7\linewidth}{!}{
  \begin{tabular}{l|ccc|cc}
    \toprule
    \multirow{2}{*}{\textbf{Methods}}  & &\textbf{mIoU}  & &\multirow{2}{*}{\textbf{Training Parmas}} & \multirow{2}{*}{\textbf{Inference Times}}\\
     &12.8m &25.6m & 51.2m   \\
    \midrule
    SSCNet  &16.32& 14.55& 10.27&-\\
    LMSCNet   &15.69& 14.13& 9.94&-\\
    MonoScene  &12.25& 12.22& 11.30&132.4M&0.281s\\
    VoxFormer  & 17.66& 16.48& 12.35&57.9M&0.307s\\
    OccFormer & 20.91& 17.90& 13.46&203.4M&0.348s\\
    HASSC& 18.98 &17.95& 13.48&-&-\\
    BRGScene & 23.27& 21.15& 15.24&161.4M&0.353s\\
    \rowcolor{gray!20}\textbf{Ours} &  \textbf{27.15}& \textbf{24.73}& \textbf{17.83}&67.2M&0.291s\\
    \bottomrule
  \end{tabular}}

\end{table}
\subsection{Main Result}

\paragraph{Quantitative Results} Table~\ref{tab:Test Quantitative Comparison} presents the quantitative comparison of existing vision-based methods on the SemanticKITTI hidden test set. Our CDScene achieves state-of-the-art performance on the SemanticKITTI. 
First, compared with BRGScene, the IoU of CDScene on the SemanticKITTI dataset is significantly improved by 2.26\%, and the mIoU is increased by 1.88\%, verifying the effectiveness of our method. The substantial improvement in IoU further confirms the advantages of the precise geometric attributes and fusion module of static cues. Meanwhile, the improvement of mIoU proves that LMMs provide rich semantic features for the model.
Second, compared with VoxFormer-T using multi-frame image input, our CDScene significantly improves SemanticKITTI, which achieves 2.39\% and 3.8\% improvements in IoU and mIoU, respectively. 
Thanks to the rich data samples on SSCBench-KITTI-360, CDScene also demonstrates significant advantages in geometric analysis and dynamic regions over current camera-based methods. Compared with Symphonies, we achieved a significant improvement of 2.76\% in the completion metric IoU. Compared to BRGScene, we achieve leading mIou in almost all categories. As shown in Table~\ref{tab:KITTI360Test Quantitative Comparison}.
To provide a more thorough comparison, we provide additional quantitative results of SSC on the SemanticKITTI validation set in Table~\ref{tab:kittival}. The results further demonstrate the effectiveness of our approach in enhancing 3D scene perception performance. Compared to previous state-of-the-art methods, CDScene far outperforms Symphonize in terms of semantic scene understanding, where the IoU and mIoU enhancement is significant for practical applications. It proves that we are not simply lowering a certain metric to accomplish semantic scene completion.

In addition, to compare the robustness of CDScene, we performed validation on SemanticKITTI-C. At the same time, we reproduce three representative methods of only camera input: BRGScene, OccFormer and MonoScene on SemanticKITTI-C. Table~\ref{tab:robust result} shows the quantitative comparison of the four methods on SemanticKITTI-C. Our method achieves a significant lead in almost all corruption categories and severities, demonstrating the robustness of our method in various degradation environments.

Furthermore, Table~\ref{tab:diff range} shows that we provide different ranges of results on the SemanticKITTI validation set. It can be seen that our method significantly surpasses the existing methods at all three distances. 

\begin{figure*}[t]
\centering
  \includegraphics[width=\textwidth]{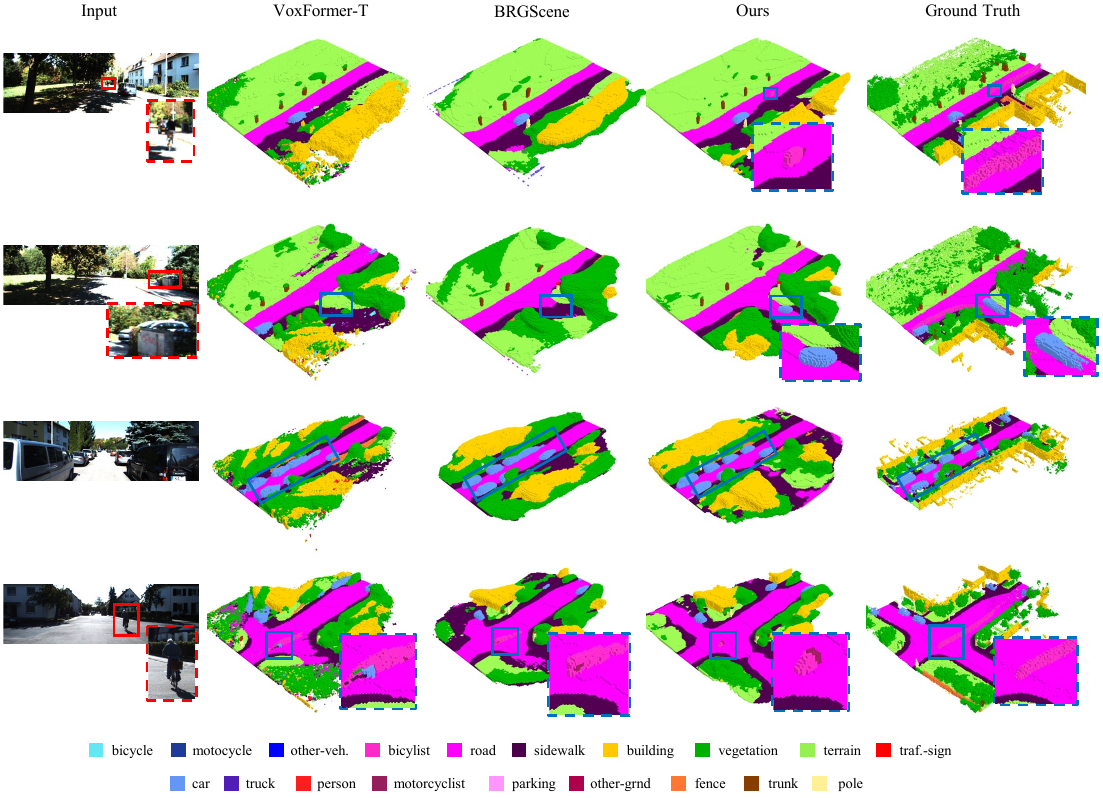}
  \caption{Qualitative results from our method and others.  The image input appears on the left, while the 3D semantic occupancy results are displayed in the following order: VoxFormer-T~\protect\cite{li2023voxformer}, BRGScene~\protect\cite{li2023stereoscene}, our CDScene and the Ground Truth.}
  \label{fig:fig6}

\end{figure*}

\paragraph{Qualitative Results}
To demonstrate our performance more intuitively, in Fig.~\ref{fig:fig1} and Fig.~\ref{fig:fig6}, we compare the results obtained from VoxFormer-T, BRGScene and CDScene on the SemanticKITTI validation set qualitative results.
Compared to VoxFormer-T, our method CDScene demonstrates robust geometric and semantic reconstruction of dynamic regions. For example, it effectively reduces long tail traces of moving vehicles and accurately identifies distant moving target objects (cyclists). This demonstrates that our method can effectively handle dynamic regions.
Furthermore, CDScene can capture better overall scene layout and produce denser semantic scene completion results than BRGScene. Our method accurately segments boundaries and avoids misclassified semantic categories in structured cars, trucks, sidewalks, and roads. In addition, CDScene has more advantages in identifying road and vehicle elements, as evidenced by the precise edge representation and geometric shape of the car in the last row of images (blue box) in Fig.~\ref{fig:fig6}. This is due to the rich geometric features provided by static cues.

\begin{figure*}[t]
\centering
  \includegraphics[width=\textwidth]{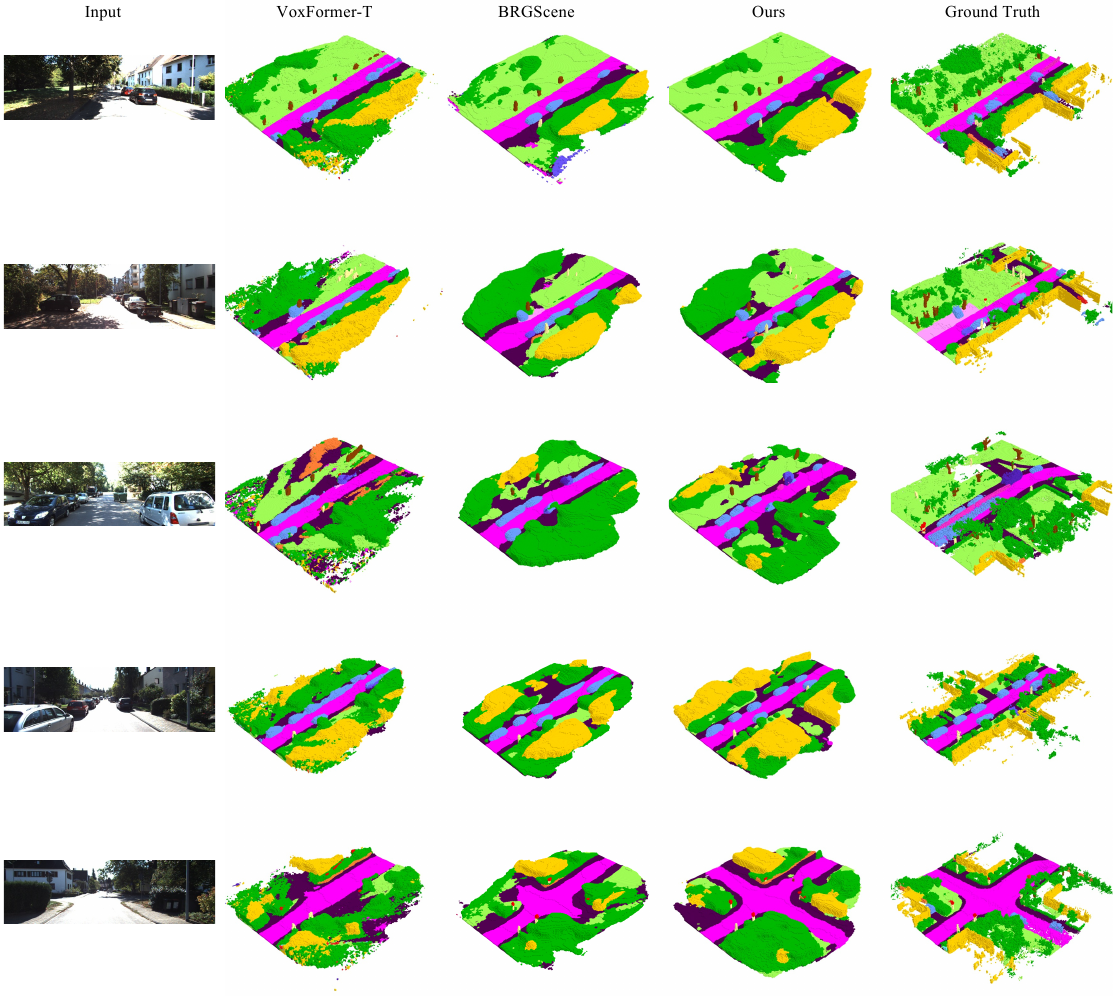}
  \caption{More Qualitative results from our method and others.}
  \label{fig:fig7}
\end{figure*}

We report the performance of more visual results on the SemanticKITTI validation set in Fig.~\ref{fig:fig7}. Overall, our method shows excellent performance for moving objects. From the details, our method segments the scene more fine-grained and maintains clear segmentation boundaries. For example, in the segmentation completion results of cars, we predict that each car has obvious separation. In contrast, other methods exhibit continuous semantic errors for occluded cars. Finally, it is surprising that our method can show excellent performance scene hallucinations in areas outside the camera's field of view.

\subsection{Ablation Studies}

\begin{table}[t]
\centering
  \caption{Ablation study for Architecture Components.}
  \label{tab:architecture}
\setlength{\tabcolsep}{3pt}
\resizebox{0.7\textwidth}{!}{
  \begin{tabular}{c|cccc|ccc}
    \toprule
     \textbf{Method}&LMMs& {Dyna} & {Stat} & {DSAF} & \textbf{IoU} & \textbf{mIoU}&\textbf{Training Params} \\

    \midrule
      Baseline&{}& {\checkmark} & {} & {} &  {42.34} & {15.32}&47.4M \\
     (a)&{\checkmark}& {\checkmark} & {} & {} &  {43.44} & {15.98}& 50.4M\\
     (b)&{\checkmark}& {} & {\checkmark} & {} &  {44.58} & {16.21}&53.2M\\
     (c)&{\checkmark}& {\checkmark} & {\checkmark} &  &  {44.75} & {16.56}&63.2M\\
     (d)&{}& {\checkmark} & {\checkmark} & \checkmark &  {44.15} & {16.38} &66.8M\\
     \rowcolor{gray!20}(e)&{\checkmark}&{\checkmark } & {\checkmark} & {\checkmark}& {\textbf{45.11}} &{\textbf{17.83}}&67.2M \\
    \bottomrule
\end{tabular}}

\end{table}

\begin{table}[ht]
  \centering
    \caption{Ablation study of static branch frame numbers.}
    \resizebox{0.3\textwidth}{!}{
  \begin{tabular}{l|rr}
    \toprule
    {\textbf{Frames}} & \textbf{IoU} & {\textbf{mIoU}} \\
    \midrule
    2 & {44.13} & {16.44} \\
    3 & {44.36} & {16.56} \\
    4 & {45.03} & {17.25} \\
   \rowcolor{gray!20}5 & {\textbf{45.11}} &{\textbf{17.83}} \\
    6 & {44.33} & {17.39} \\
    7 & {44.98} & {16.85} \\
    8 & {44.53} & {16.74} \\
    \bottomrule
  \end{tabular}
}
  \label{tab:frames}
\end{table}
\begin{table}[t!]
  \centering
    \caption{Ablation study for Dynamic-Static Adaptive Fusion Module.}
    \resizebox{0.5\textwidth}{!}{
  \begin{tabular}{l|rr}
    \toprule
    {\textbf{Setting}} & {\textbf{IoU}} & {\textbf{mIoU}} \\
    \midrule
    {Add-Conv Fusion} & {43.14} & {15.44} \\
    {Cat-Conv Fusion} & {44.46} & {15.88} \\
    {Attention Fusion} & {44.70} & {16.32} \\
    \rowcolor{gray!20}Dynamic-Static Adaptive Fusion & {\textbf{45.11}} &{\textbf{17.83}} \\
    \bottomrule
  \end{tabular}}

  \label{tab:evc}
\end{table}
\paragraph{Ablation on the Architectural Components}  Table~\ref{tab:architecture} shows the breakdown of various architectural components in CDScene, including large multimodal models (LMMs), dynamic branch (Dyna), static branch (Stat), and dynamic-static adaptive fusion (DSAF). We will analyze the contents of Table~\ref{tab:architecture} line by line. 
\begin{description}
    \item[Baseline:] First, we use RepVit as the image encoder and sparse 3D convolutional network as the voxel encoder, following BRGStereo’s view transformation paradigm as a baseline. 
    \item[(a):] When adding visual language features, the model obtains rich semantic features and mIoU is significantly improved. 
    \item[(b):] The introduction of the static branch brings dense static features and rich geometric attributes, increasing mIoU and IoU by 0.66\% and 1.1\% respectively.
    \item[(c):] When both dynamic branch and static branch are used for convolutional channel fusion, mIoU and IoU are improved by 1.24\% and 2.41\% respectively.
    \item[(d):] Setting (d) further illustrates the significant performance degradation when the full model lacks explicit semantic features.
    \item[(e):] The final full model achieves 2.77\% IoU and 2.51\% mIoU improvement over the baseline, demonstrating that all these components contribute to the best results.
\end{description}
\paragraph{Ablation on the Static Branch Frame Numbers} Table~\ref{tab:frames} shows how the model's performance initially improves with an increase in the number of frames used as temporal input, but later declines due to inaccuracies in camera-extrinsic matrix alignment between historical frames and the current system. This can disrupt the extraction of voxel features and cause 3D points to map erroneously onto image features from past frames.
\paragraph{Ablation on the Dynamic-Static Adaptive Fusion}
To deeply study the functions of the dynamic-static adaptive fusion module, we compared it with several mainstream fusion methods, shown in Table~\ref{tab:evc}. First, for add-conv, we found that using element-wise addition and performing convolution operations did not improve metrics. We analyze that the feature maps from the two branches are not completely aligned, and a simple add-conv operation may damage the original feature distribution. Secondly, using cat-conv to concatenate the channel numbers of two features and perform convolution can bring a small improvement. Simple Attention fusion can enable dynamic features to globally query static features, which brings better results but increases the consumption of computing resources. Finally, based on the analysis of static geometric clues and dynamic structure clues, our dynamic-static adaptive fusion module adaptively integrates both advantages and significantly improves mIoU and IoU, respectively.

\subsection{Discussions} 
CDScene is explored for dynamic scenes and shows strong performance in benchmarks. However, its running speed can be further improved. Also, the SSC task in multi-camera settings needs to be explored.
{Although our proposed CDScene method is designed within a modular perception framework, it can also provide valuable benefits to end-to-end autonomous driving pipelines. The generated semantically complete 3D scene not only enhances environmental awareness but also offers interpretable and structured spatial information that can be used to supervise or regularize driving policies. For instance, recent studies have shown that integrating semantic occupancy maps or voxelized scene context into end-to-end driving networks improves safety and generalization. In future work, we plan to explore the integration of CDScene into closed-loop driving systems, where scene completion can serve as a robust spatial prior or be jointly optimized with policy learning tasks.} 
\section{Conclusion}
In this paper, we propose a novel method, CDScene.
First, we leverage a large multi-modal model to extract 2D explicit semantics.
Second, we exploit the features of monocular and stereo depth to decouple scene information into dynamic and static features. The dynamic features contain structural relationships around dynamic objects, and the static features contain dense contextual spatial information.
Finally, we design a dynamic-static adaptive fusion module to effectively extract and aggregate complementary features, achieving robust and accurate semantic scene completion in autonomous driving scenarios. Extensive experimental results on the SemanticKITTI, SSCBench-KITTI360, and SemanticKITTI-C datasets demonstrate the superiority and robustness of CDScene over existing state-of-the-art methods.

\section{Acknowledgments}
This work was supported by the National Natural Science Foundation of China (Grant Nos.62225205) and Postgraduate Scientific Research Innovation Project of Hunan Province (CX20240427).









\bibliographystyle{elsarticle-num} 
\bibliography{ref}

\begin{thebibliography}{10}
\expandafter\ifx\csname url\endcsname\relax
  \def\url#1{\texttt{#1}}\fi
\expandafter\ifx\csname urlprefix\endcsname\relax\def\urlprefix{URL }\fi
\expandafter\ifx\csname href\endcsname\relax
  \def\href#1#2{#2} \def\path#1{#1}\fi

\bibitem{shen2023flowformer}
Y.~Shen, L.~Hui, Flowformer: 3d scene flow estimation for point clouds with transformers, Knowledge-Based Systems 280 (2023) 111041.

\bibitem{song2017semantic}
S.~Song, F.~Yu, A.~Zeng, A.~X. Chang, M.~Savva, T.~Funkhouser, Semantic scene completion from a single depth image, in: CVPR, 2017, pp. 1746--1754.

\bibitem{yan2021sparse}
X.~Yan, J.~Gao, J.~Li, R.~Zhang, Z.~Li, R.~Huang, S.~Cui, Sparse single sweep lidar point cloud segmentation via learning contextual shape priors from scene completion, in: AAAI, Vol.~35, 2021, pp. 3101--3109.

\bibitem{guo2018view}
Y.~Guo, X.~Tong, View-volume network for semantic scene completion from a single depth image, in: IJCAI, 2018, pp. 726--732.

\bibitem{roldao2020lmscnet}
L.~Roldao, R.~de~Charette, A.~Verroust-Blondet, Lmscnet: Lightweight multiscale 3d semantic completion, in: 3DV, IEEE, 2020, pp. 111--119.

\bibitem{cao2022monoscene}
A.-Q. Cao, R.~de~Charette, Monoscene: Monocular 3d semantic scene completion, in: CVPR, 2022.

\bibitem{li2023voxformer}
Y.~Li, Z.~Yu, C.~Choy, C.~Xiao, J.~M. Alvarez, S.~Fidler, C.~Feng, A.~Anandkumar, Voxformer: Sparse voxel transformer for camera-based 3d semantic scene completion, in: CVPR, 2023.

\bibitem{li2023stereoscene}
B.~Li, Y.~Sun, X.~Jin, W.~Zeng, Z.~Zhu, X.~Wang, Y.~Zhang, J.~Okae, H.~Xiao, D.~Du, Stereoscene: Bev-assisted stereo matching empowers 3d semantic scene completion, arXiv preprint arXiv:2303.13959 (2023).

\bibitem{jiang2024symphonize}
H.~Jiang, T.~Cheng, N.~Gao, H.~Zhang, T.~Lin, W.~Liu, X.~Wang, Symphonize 3d semantic scene completion with contextual instance queries, in: Proceedings of the IEEE/CVF Conference on Computer Vision and Pattern Recognition, 2024, pp. 20258--20267.

\bibitem{tits2024instance}
H.~Xiao, H.~Xu, W.~Kang, Y.~Li, Instance-aware monocular 3d semantic scene completion, IEEE Transactions on Intelligent Transportation Systems (2024).

\bibitem{mixssc}
M.~Wang, Y.~Ding, Y.~Liu, Y.~Qin, R.~Li, Z.~Tang, Mixssc: Forward-backward mixture for vision-based 3d semantic scene completion, IEEE Transactions on Circuits and Systems for Video Technology (2025) 1--1.

\bibitem{wang2025vlscene}
M.~Wang, H.~Pi, R.~Li, Y.~Qin, Z.~Tang, K.~Li, Vlscene: Vision-language guidance distillation for camera-based 3d semantic scene completion, in: Proceedings of the AAAI Conference on Artificial Intelligence, Vol.~39, 2025, pp. 7808--7816.

\bibitem{bhat2021adabins}
S.~F. Bhat, I.~Alhashim, P.~Wonka, Adabins: Depth estimation using adaptive bins, in: Proceedings of the IEEE/CVF conference on computer vision and pattern recognition, 2021, pp. 4009--4018.

\bibitem{yin2022towards}
W.~Yin, J.~Zhang, O.~Wang, S.~Niklaus, S.~Chen, Y.~Liu, C.~Shen, Towards accurate reconstruction of 3d scene shape from a single monocular image, IEEE Transactions on Pattern Analysis and Machine Intelligence 45~(5) (2022) 6480--6494.

\bibitem{yan2023dsc}
W.~Yan, L.~Dong, W.~Ma, Q.~Mi, H.~Zha, Dsc-mde: Dual structural contexts for monocular depth estimation, Knowledge-Based Systems 263 (2023) 110308.

\bibitem{gu2020cascade}
X.~Gu, Z.~Fan, S.~Zhu, Z.~Dai, F.~Tan, P.~Tan, Cascade cost volume for high-resolution multi-view stereo and stereo matching, in: Proceedings of the IEEE/CVF conference on computer vision and pattern recognition, 2020, pp. 2495--2504.

\bibitem{yao2018mvsnet}
Y.~Yao, Z.~Luo, S.~Li, T.~Fang, L.~Quan, Mvsnet: Depth inference for unstructured multi-view stereo, in: Proceedings of the European conference on computer vision (ECCV), 2018, pp. 767--783.

\bibitem{Yu_2020_fastmvsnet}
Z.~Yu, S.~Gao, Fast-mvsnet: Sparse-to-dense multi-view stereo with learned propagation and gauss-newton refinement, in: Proceedings of the IEEE/CVF Conference on Computer Vision and Pattern Recognition (CVPR), 2020.

\bibitem{chuah2022semantic}
W.~Chuah, R.~Tennakoon, R.~Hoseinnezhad, D.~Suter, A.~Bab-Hadiashar, Semantic guided long range stereo depth estimation for safer autonomous vehicle applications, IEEE Transactions on Intelligent Transportation Systems 23~(10) (2022) 18916--18926.

\bibitem{behley2019semantickitti}
J.~Behley, M.~Garbade, A.~Milioto, J.~Quenzel, S.~Behnke, C.~Stachniss, J.~Gall, Semantickitti: A dataset for semantic scene understanding of lidar sequences, in: ICCV, 2019, pp. 9297--9307.

\bibitem{li2023sscbench}
Y.~Li, S.~Li, X.~Liu, M.~Gong, K.~Li, N.~Chen, Z.~Wang, Z.~Li, T.~Jiang, F.~Yu, Y.~Wang, H.~Zhao, Z.~Yu, C.~Feng, Sscbench: Monocular 3d semantic scene completion benchmark in street views, arXiv preprint arXiv:2306.09001 (2023).

\bibitem{zou2021udnet}
H.~Zou, X.~Yang, T.~Huang, C.~Zhang, Y.~Liu, W.~Li, F.~Wen, H.~Zhang, Up-to-down network: Fusing multi-scale context for 3d semantic scene completion, in: 2021 IEEE/RSJ International Conference on Intelligent Robots and Systems (IROS), IEEE, 2021, pp. 16--23.

\bibitem{zhang2018efficient}
J.~Zhang, H.~Zhao, A.~Yao, Y.~Chen, L.~Zhang, H.~Liao, Efficient semantic scene completion network with spatial group convolution, in: ECCV, 2018, pp. 733--749.

\bibitem{rist2021semantic}
C.~B. Rist, D.~Emmerichs, M.~Enzweiler, D.~M. Gavrila, Semantic scene completion using local deep implicit functions on lidar data, TPAMI 44~(10) (2021) 7205--7218.

\bibitem{cai2021semantic}
Y.~Cai, X.~Chen, C.~Zhang, K.-Y. Lin, X.~Wang, H.~Li, Semantic scene completion via integrating instances and scene in-the-loop, in: CVPR, 2021, pp. 324--333.

\bibitem{li2019rgbd}
J.~Li, Y.~Liu, D.~Gong, Q.~Shi, X.~Yuan, C.~Zhao, I.~Reid, Rgbd based dimensional decomposition residual network for 3d semantic scene completion, in: CVPR, 2019, pp. 7693--7702.

\bibitem{wan2024bi}
Y.~Wan, P.~Lv, L.~Sun, Y.~Yang, J.~Hao, Bi-interfusion: A bidirectional cross-fusion framework with semantic-guided transformers in lidar-camera fusion, Knowledge-Based Systems 305 (2024) 112577.

\bibitem{yang2021ssasc}
X.~Yang, H.~Zou, X.~Kong, T.~Huang, Y.~Liu, W.~Li, F.~Wen, H.~Zhang, Semantic segmentation-assisted scene completion for lidar point clouds, in: 2021 IEEE/RSJ International Conference on Intelligent Robots and Systems (IROS), IEEE, 2021, pp. 3555--3562.

\bibitem{mei2023sscrs}
J.~Mei, Y.~Yang, M.~Wang, T.~Huang, X.~Yang, Y.~Liu, Ssc-rs: Elevate lidar semantic scene completion with representation separation and bev fusion, in: 2023 IEEE/RSJ International Conference on Intelligent Robots and Systems (IROS), IEEE, 2023, pp. 1--8.

\bibitem{huang2023tri}
Y.~Huang, W.~Zheng, Y.~Zhang, J.~Zhou, J.~Lu, Tri-perspective view for vision-based 3d semantic occupancy prediction, in: CVPR, 2023, pp. 9223--9232.

\bibitem{zhang2023occformer}
Y.~Zhang, Z.~Zhu, D.~Du, Occformer: Dual-path transformer for vision-based 3d semantic occupancy prediction, arXiv preprint arXiv:2304.05316 (2023).

\bibitem{philion2020lift}
J.~Philion, S.~Fidler, Lift, splat, shoot: Encoding images from arbitrary camera rigs by implicitly unprojecting to 3d, in: ECCV, Springer, 2020, pp. 194--210.

\bibitem{wei2023surroundocc}
Y.~Wei, L.~Zhao, W.~Zheng, Z.~Zhu, J.~Zhou, J.~Lu, Surroundocc: Multi-camera 3d occupancy prediction for autonomous driving, in: ICCV, 2023, pp. 21729--21740.

\bibitem{yao2023ndc}
J.~Yao, C.~Li, K.~Sun, Y.~Cai, H.~Li, W.~Ouyang, H.~Li, Ndc-scene: Boost monocular 3d semantic scene completion in normalized device coordinates space, in: ICCV, 2023, pp. 9455--9465.

\bibitem{zheng2024monoocc}
Y.~Zheng, X.~Li, P.~Li, Y.~Zheng, B.~Jin, C.~Zhong, X.~Long, H.~Zhao, Q.~Zhang, Monoocc: Digging into monocular semantic occupancy prediction, arXiv preprint arXiv:2403.08766 (2024).

\bibitem{wang2024h2gformer}
Y.~Wang, C.~Tong, H2gformer: Horizontal-to-global voxel transformer for 3d semantic scene completion, in: Proceedings of the AAAI Conference on Artificial Intelligence, Vol.~38, 2024, pp. 5722--5730.

\bibitem{wang2024HASSC}
S.~Wang, J.~Yu, W.~Li, W.~Liu, X.~Liu, J.~Chen, J.~Zhu, Not all voxels are equal: Hardness-aware semantic scene completion with self-distillation, in: Proceedings of the IEEE/CVF Conference on Computer Vision and Pattern Recognition, 2024, pp. 14792--14801.

\bibitem{li2023bevstereo}
Y.~Li, H.~Bao, Z.~Ge, J.~Yang, J.~Sun, Z.~Li, Bevstereo: Enhancing depth estimation in multi-view 3d object detection with temporal stereo, in: Proceedings of the AAAI Conference on Artificial Intelligence, Vol.~37, 2023, pp. 1486--1494.

\bibitem{Li_2023_CVPR}
R.~Li, D.~Gong, W.~Yin, H.~Chen, Y.~Zhu, K.~Wang, X.~Chen, J.~Sun, Y.~Zhang, Learning to fuse monocular and multi-view cues for multi-frame depth estimation in dynamic scenes, in: Proceedings of the IEEE/CVF Conference on Computer Vision and Pattern Recognition (CVPR), 2023, pp. 21539--21548.

\bibitem{alayrac2022flamingo}
J.-B. Alayrac, J.~Donahue, P.~Luc, A.~Miech, I.~Barr, Y.~Hasson, K.~Lenc, A.~Mensch, K.~Millican, M.~Reynolds, et~al., Flamingo: a visual language model for few-shot learning, Advances in neural information processing systems 35 (2022) 23716--23736.

\bibitem{li2023blip}
J.~Li, D.~Li, S.~Savarese, S.~Hoi, Blip-2: Bootstrapping language-image pre-training with frozen image encoders and large language models, in: International conference on machine learning, PMLR, 2023, pp. 19730--19742.

\bibitem{hasan2024vision}
M.~Z. Hasan, J.~Chen, J.~Wang, M.~S. Rahman, A.~Joshi, S.~Velipasalar, C.~Hegde, A.~Sharma, S.~Sarkar, Vision-language models can identify distracted driver behavior from naturalistic videos, IEEE Transactions on Intelligent Transportation Systems (2024).

\bibitem{zhu2023minigpt}
D.~Zhu, J.~Chen, X.~Shen, X.~Li, M.~Elhoseiny, Minigpt-4: Enhancing vision-language understanding with advanced large language models, arXiv preprint arXiv:2304.10592 (2023).

\bibitem{NEURIPS2023InstructBLIP}
W.~Dai, J.~Li, D.~LI, A.~Tiong, J.~Zhao, W.~Wang, B.~Li, P.~N. Fung, S.~Hoi, Instructblip: Towards general-purpose vision-language models with instruction tuning, in: A.~Oh, T.~Naumann, A.~Globerson, K.~Saenko, M.~Hardt, S.~Levine (Eds.), Advances in Neural Information Processing Systems, Vol.~36, Curran Associates, Inc., 2023, pp. 49250--49267.

\bibitem{radford2021clip}
A.~Radford, J.~W. Kim, C.~Hallacy, A.~Ramesh, G.~Goh, S.~Agarwal, G.~Sastry, A.~Askell, P.~Mishkin, J.~Clark, et~al., Learning transferable visual models from natural language supervision, in: International conference on machine learning, PMLR, 2021, pp. 8748--8763.

\bibitem{li2022lseg}
B.~Li, K.~Q. Weinberger, S.~Belongie, V.~Koltun, R.~Ranftl, Language-driven semantic segmentation, arXiv preprint arXiv:2201.03546 (2022).

\bibitem{ghiasi2022openseg}
G.~Ghiasi, X.~Gu, Y.~Cui, T.-Y. Lin, Scaling open-vocabulary image segmentation with image-level labels, in: European Conference on Computer Vision, Springer, 2022, pp. 540--557.

\bibitem{dong2023maskclip}
X.~Dong, J.~Bao, Y.~Zheng, T.~Zhang, D.~Chen, H.~Yang, M.~Zeng, W.~Zhang, L.~Yuan, D.~Chen, et~al., Maskclip: Masked self-distillation advances contrastive language-image pretraining, in: Proceedings of the IEEE/CVF Conference on Computer Vision and Pattern Recognition, 2023, pp. 10995--11005.

\bibitem{chen2023clip2scene}
R.~Chen, Y.~Liu, L.~Kong, X.~Zhu, Y.~Ma, Y.~Li, Y.~Hou, Y.~Qiao, W.~Wang, Clip2scene: Towards label-efficient 3d scene understanding by clip, in: Proceedings of the IEEE/CVF Conference on Computer Vision and Pattern Recognition, 2023, pp. 7020--7030.

\bibitem{Peng2023OpenScene}
S.~Peng, K.~Genova, C.~M. Jiang, A.~Tagliasacchi, M.~Pollefeys, T.~Funkhouser, Openscene: 3d scene understanding with open vocabularies, 2023.

\bibitem{kuo2023F-VLM}
W.~Kuo, Y.~Cui, X.~Gu, A.~Piergiovanni, A.~Angelova, Open-vocabulary object detection upon frozen vision and language models, in: The Eleventh International Conference on Learning Representations, 2023.

\bibitem{he2016resnet}
K.~He, X.~Zhang, S.~Ren, J.~Sun, Deep residual learning for image recognition, in: CVPR, 2016, pp. 770--778.

\bibitem{dosovitskiy2020vit}
A.~Dosovitskiy, L.~Beyer, A.~Kolesnikov, D.~Weissenborn, X.~Zhai, T.~Unterthiner, M.~Dehghani, M.~Minderer, G.~Heigold, S.~Gelly, et~al., An image is worth 16x16 words: Transformers for image recognition at scale, arXiv preprint arXiv:2010.11929 (2020).

\bibitem{vaswani2017attention}
A.~Vaswani, N.~Shazeer, N.~Parmar, J.~Uszkoreit, L.~Jones, A.~N. Gomez, {\L}.~Kaiser, I.~Polosukhin, Attention is all you need, Advances in neural information processing systems 30 (2017).

\bibitem{wang2023repvit}
A.~Wang, H.~Chen, Z.~Lin, J.~Han, G.~Ding, Repvit: Revisiting mobile cnn from vit perspective (2023).
\newblock \href {http://arxiv.org/abs/2307.09283} {\path{arXiv:2307.09283}}.

\bibitem{guo2019gwcnet}
X.~Guo, K.~Yang, W.~Yang, X.~Wang, H.~Li, Group-wise correlation stereo network, in: Proceedings of the IEEE/CVF conference on computer vision and pattern recognition, 2019, pp. 3273--3282.

\bibitem{Hassani_2023_Neighborhood}
A.~Hassani, S.~Walton, J.~Li, S.~Li, H.~Shi, Neighborhood attention transformer, in: Proceedings of the IEEE/CVF Conference on Computer Vision and Pattern Recognition (CVPR), 2023, pp. 6185--6194.

\bibitem{chen2017aspp}
L.-C. Chen, G.~Papandreou, I.~Kokkinos, K.~Murphy, A.~L. Yuille, Deeplab: Semantic image segmentation with deep convolutional nets, atrous convolution, and fully connected crfs, IEEE transactions on pattern analysis and machine intelligence 40~(4) (2017) 834--848.

\bibitem{li2023bevdepth}
Y.~Li, Z.~Ge, G.~Yu, J.~Yang, Z.~Wang, Y.~Shi, J.~Sun, Z.~Li, Bevdepth: Acquisition of reliable depth for multi-view 3d object detection, in: AAAI, Vol.~37, 2023, pp. 1477--1485.

\bibitem{Geiger2012kitti}
A.~Geiger, P.~Lenz, R.~Urtasun, Are we ready for autonomous driving? the kitti vision benchmark suite, in: CVPR, 2012.

\bibitem{Liao2022kitti360}
Y.~Liao, J.~Xie, A.~Geiger, {KITTI}-360: A novel dataset and benchmarks for urban scene understanding in 2d and 3d, TPAMI (2022).

\bibitem{kong2023robodepth}
L.~Kong, S.~Xie, H.~Hu, L.~X. Ng, B.~R. Cottereau, W.~T. Ooi, Robodepth: Robust out-of-distribution depth estimation under corruptions, in: Advances in Neural Information Processing Systems, 2023.

\bibitem{lin2017feature}
T.-Y. Lin, P.~Doll{\'a}r, R.~Girshick, K.~He, B.~Hariharan, S.~Belongie, Feature pyramid networks for object detection, in: CVPR, 2017, pp. 2117--2125.

\end{thebibliography}
\end{document}